\documentclass{article}
\usepackage{arxiv}

\usepackage{times}
\usepackage{soul}
\usepackage{url}
\usepackage[hidelinks]{hyperref}
\usepackage[utf8]{inputenc}
\usepackage[small]{caption}
\usepackage{graphicx}
\usepackage{amsmath}
\usepackage{amsthm}
\usepackage{booktabs}
\usepackage{algorithm}
\usepackage{xcolor}

\title{Social Learning through Interactions with Other Agents: A Survey}

\author{
Dylan Hillier$^{1,2}$
\And
Cheston Tan$^{1}$\And
Jing Jiang$^{2}$\\
$^1$Centre for Frontier AI Research, Institute of High Performance Computing, A*STAR\\
$^2$Singapore Management University\\
das.hillier.2023@phdcs.smu.edu.sg,
cheston\_tan@cfar.a-star.edu.sg,
jingjiang@smu.edu.sg
}

\begin{document}
\maketitle

\begin{abstract}
    Social learning plays an important role in the development of human intelligence. As children, we imitate our parents' speech patterns until we are able to produce sounds; we learn from them praising us and scolding us; and as adults, we learn by working with others. In this work, we survey the degree to which this paradigm -- \textit{social learning} -- has been mirrored in machine learning. In particular, since learning socially requires interacting with others, we are interested in how \textit{embodied agents} can and have utilised these techniques. This is especially in light of the degree to which recent advances in natural language processing (NLP) enable us to perform new forms of social learning. We look at how behavioural cloning and next-token prediction mirror human imitation, how learning from human feedback mirrors human education, and how we can go further to enable fully communicative agents that learn from each other.
We find that while individual social learning techniques have been used successfully, there has been little unifying work showing how to bring them together into socially embodied agents.
\end{abstract}

\section{Introduction}
Social learning is ``learning that is influenced by the observation of another individual or their products''~\cite{bonnefon2023machine}. The ``embodiment hypothesis'' in cognitive science suggests that intelligence emerges as a result of embodiment within a social environment, that ``starting as a baby grounded in a physical, social, and linguistic world is crucial to the development of the flexible and inventive intelligence that characterizes humankind''~\cite{smith2005development}. We can define this embodiment as being when ``the environment is acting upon the individual and the individual is acting upon the environment''~\cite{bolotta2022social}. We explore how social embodiment -- learning that is influenced by the actions and products of other agents in the learning environment -- has played a role in the development of AI, and how it could influence the development of embodied agents in the real world.

Despite often not being recognised as such, a large proportion of AI research can be considered or framed as employing aspects of social learning. In particular, models trained on human-produced data often implicitly rely on learning to imitate humans. Take for example language model pretraining, in which the model is trained to directly reproduce human text. The form of this social learning can have important implications, for example through the reproduction of human biases in textual data~\cite{bender2021dangers} and facial recognition~\cite{buolamwini2018gender}. As such, the prevalence and impact of AI systems trained in this way warrants a more thorough investigation of the mechanics of social learning.

From another perspective, social learning techniques have largely seen success on supervised offline learning tasks with fixed datasets. Real-world deployment often requires accounting for more complicated, two-way interactions between agent and environment. This has traditionally been done through individualistic reinforcement learning methods, which have issues like sample inefficiency, unspecified reward functions, and poor explainability~\cite{dulac2021challenges}. In contrast, humans -- who develop in a socially embodied setting -- are able to quickly adapt to changes in the environment. Since communication between agents has been made easier by the development of LLMs, it is an apt time to research the development of agents using social learning.

\subsection{Social Learning in Humans}

Social learning plays an instrumental role in the way humans learn~\cite{tomasello2009cultural}. We do not learn individualistically by simply observing the world -- instead, we take advantage of the fact that other humans have learnt before us and have built up cultural knowledge.
Indeed, \textit{Homo Sapiens'} exploitation of cultural knowledge is exactly what has allowed us to build complex societies and dominate the globe. Rather than having bigger brains (and thereby having a cognitive advantage), our ability to ``accumulate information across generations and develop well-adapted tools, beliefs, and practices that no individual could invent on their own'' is responsible for our success~\cite{boyd2011cultural}.

Similarly, the transmission and generation of this cultural knowledge is of paramount importance in human development at both an individual and societal level. Social learning enables this cultural transmission and development through our interactions with others. This suggests to us that the promising route to achieving more advanced agents lies not just in building bigger artificial brains (scaling) but also in building artificial brains socially, that can learn from and teach other agents, and take advantage of and develop this rich cultural knowledge base~\cite{bolotta2022social}. 

\begin{figure}[t]
    \centering
    \includegraphics[scale=0.78]{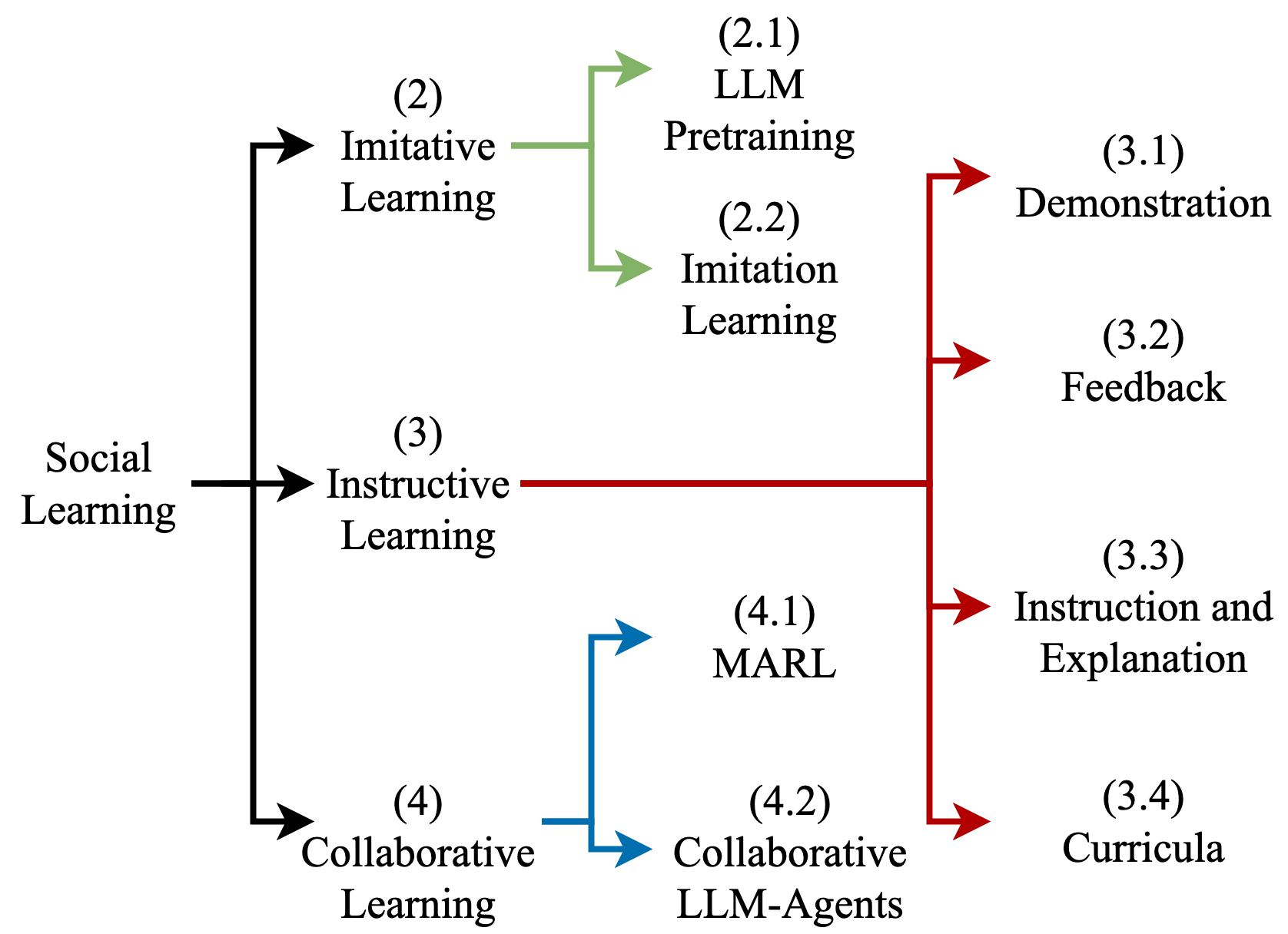}
    \caption{The sections of this survey structured in relation to Tomasello's classification of Social Learning.}
    \label{fig:section diagram}
\end{figure}

According to Tomasello's influential theory of Social Learning, it can be divided into Imitative Learning, Instructive Learning, and Collaborative Learning~\cite{tomasello1993cultural}.
The organisation of this survey follows this division, as depicted in Figure~\ref{fig:section diagram}.

\paragraph{Imitative Learning.}
In Tomasello's theory of Social Learning, children learn by copying adults in a number of ways. A distinction is drawn between \textit{emulation} and \textit{true imitation}. In the case of the former, the child attempts to reproduce the actions and outcomes of the adult demonstrator; in the latter, the child additionally models the intent behind the action and learns from that. Such \textit{true imitation} seems to require at the least a nascent theory of mind, insofar as it requires modelling the intents of others.

\paragraph{Instructive Learning.}
Instructive learning refers to the process by which an adult or expert learner tries to teach a non-expert learner. Other than explicit instruction, this process can include feedback, explanations, demonstrations and construction of a curriculum. Collectively, the process by which these mechanisms are employed to aid a learner is referred to as scaffolding~\cite{van2010scaffolding}.

\paragraph{Collaborative Learning.}
Collaborative learning is the setting in which neither or none of the collaborators are experts. In these cases, the learners must work together to progress in the learning task. 
From a developmental perspective, this emerges during early education -- in contrast to imitation learning which is observed almost from birth~\cite{tomasello1993cultural}. However, it plays an important role in childhood development, especially in the development of language outside of their asymmetric interactions with adult teachers. For example, Vygotsky and Cole~\shortcite{vygotsky1978mind} emphasise the role of childhood play in enabling peers to practice social roles within their capabilities.
\begin{figure}[t]
    \centering
    \includegraphics[scale=0.78]{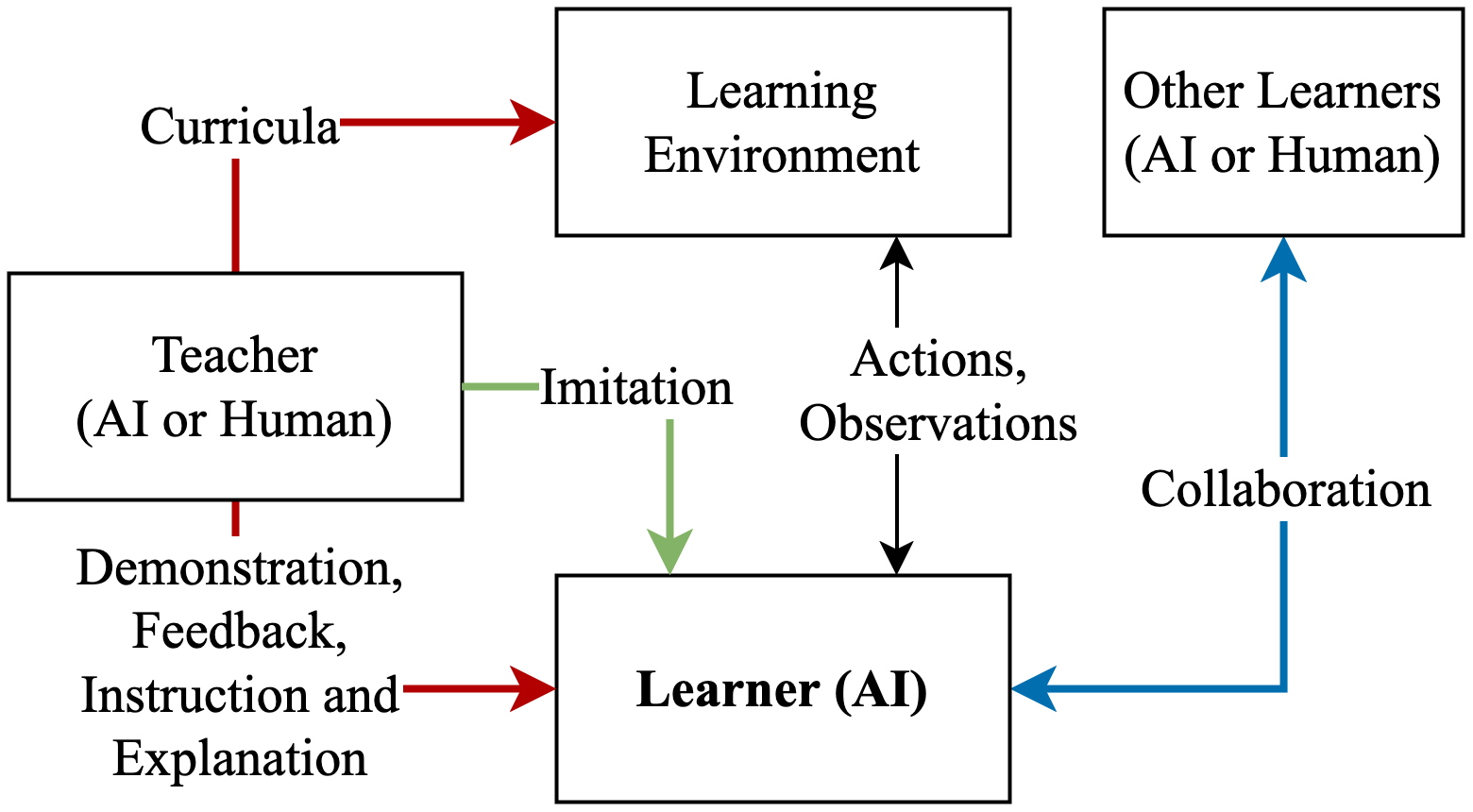}
    \caption{The relationships among different social learning techniques, agents and the learning environment. Note: 
    \textit{Imitation} is initiated by the learner rather than the teacher. As in Figure~\ref{fig:section diagram}, the learning approaches are coloured according to their classification in Tomasello's account of Social Learning.}
    \label{fig:Learning Strategies}
\end{figure}
\subsection{Social Learning in AI}
\label{sec: social_learn_intro}
At first glance, some forms of social learning techniques have been widely used in AI at large~\cite{duenez2023social}.
Diving deeper, we seek to make sense of these from a unified perspective, and examine whether there are more aspects of social learning that can be applied to AI.

In the simplest case, training data is generated or curated by humans. Furthermore, many models are trained to generate more of this human-supplied data -- this often consitutes a form of imitation learning. For example, a generative image model may be trained to imitate the art or photography of humans. While such imitative models are powerful, they are also inherently limited to recreating the level of the source of their learning~\cite{yiu2023transmission}. Several instructive techniques like feedback~\cite{ouyang2022training} and curricula~\cite{bengio2009curriculum} are also widely used in various domains. Finally, collaborative learning regimes have been explored in the Multi Agent Reinforcement Learning (MARL) literature~\cite{gronauer2022multi}. All these techniques and their relations are depicted in Figure~\ref{fig:Learning Strategies}.
\subsection{LLM Agents}
\label{llm_agent_intro}
The development of LLMs poses a striking opportunity to enable embodied agents to learn socially. Indeed, early experiments with creating embodied agents almost entirely from LLMs have shown that such agents are able to achieve remarkable zero-shot performance on embodied tasks~\cite{wang2024voyager}. This has been boosted by the development of Vision Language Models (VLMs) and other multimodal models that have been used to create agents that can naturally leverage human demonstration data and cultural knowledge~\cite{zitkovich2023rt,wang2023jarvis}. In addition, since they are language models, such agents possess an inherent ability to communicate with each other, without the need to develop such methods from scratch. This has accordingly enabled the creation of cooperative groups of agents~\cite{chen2024agentverse}.

\subsection{Objective of This Survey}

There has not been a recent survey on social learning covering the developments arising from large language models (LLMs). Meanwhile, a growing number of works have called for more exploration of social learning approaches~\cite{bonnefon2023machine,duenez2023social,bolotta2022social,yiu2023transmission}.
In this survey, we aim to provide a review of recent works that use social learning techniques with a view to building agents that can learn just through interacting with others. Given the relative maturity of imitation techniques for training generative models, in this survey we put more focus on the potential for training agents using instructive and  collaborative forms of social learning, as this has been most directly enabled by the emergence of LLM Agents.

As depicted in Figure~\ref{fig:section diagram}, we first consider existing approaches to imitation learning, both in the wider machine learning literature, as well as in their application to building embodied agents. Relatedly, we briefly consider some of the work that has explored how to enable this imitation learning in Section~\ref{sec: demo}. In Section~\ref{sec: fed}, we then extensively explore how feedback has been used for aligning LLMs, and how richer forms of feedback might be leveraged. Many of these lessons are also applicable to Explanations and Instructions as briefly covered in Section~\ref{sec: i&e}. We then explore the degree to which curricula are useful for these embodied model given some negative results in Section~\ref{sec: cur}. We cover how MARL has traditionally been used for collaborative learning in Section~\ref{sec: marl} Finally, we cover the recent development of collaborative LLM-agents in Section~\ref{sec: llm col}.

More generally, by carefully considering the gaps between human social learning paradigms and current techniques used in training embodied agents we hope to inspire research that more fully utilises the advantages offered by social and cultural learning.

\section{Imitative Learning}
\label{sec: im}
\subsection{LLM Pretraining}
Large language models are largely trained on the task of \textit{next token prediction} -- essentially predicting the next word in a sentence. This is arguably a form of imitative learning in so far as the model is fed observations of the productions of a human agent (the sequence of tokens), and learns to autoregressively predict that sequence of tokens. As indicated in Section~\ref{llm_agent_intro}, such models have achieved remarkable feats, to the point of matching humans on a wide number of tasks.

Nevertheless this acquisition process is highly data hungry -- GPT-3 was trained on over 200 billion tokens, whereas the average 12 year old would have only seen roughly 100 million~\cite{warstadt-etal-2023-findings}. One potential explanation for this is that the techniques used for training LLMs do not consist in \textit{true imitation} in the sense described by Tomasello et al.~\shortcite{tomasello1993cultural} and rather perform ``emulation''. However, Kosinski~\shortcite{kosinski2023theory} argues that LLMs already possess theory of mind, although this position has been criticised~\cite{yiu2023transmission}.

 More pointedly, during the training regime, a large language model is devoid of the socially-embodied context that typically accompanies a child learner, such as being able to build up a model of the social environment that may be useful for prediction. In this context, language models have been shown to recreate human biases, reproduce a static view in the face of changing social mores, and lack accountability~\cite{bender2021dangers}.

We have already seen how childhood development relies on other learning regimes~\cite{tomasello1993cultural}, therefore what remains to be seen is the development of models that can actually leverage this socially embodied setting to further learn and adapt through their collaboration with other agents. Indeed, this has begun to be explored through building social simulation environments~\cite{liu2023training}. Hopefully this will enable us to move beyond LLMs just acting as ``tools of imitation''~\cite{yiu2023transmission} and mitigate some of the social issues mentioned above.

\subsection{Imitation Learning}
The field of Imitation learning itself (and indeed its dual -- instruction by demonstrations) has been a popular reward-free method of training embodied agents. We do not aim to provide a comprehensive overview of the field, as covered by e.g.~\cite{hussein2017imitation,zheng2022imitation}. Key examples of techniques in the field include:
\begin{itemize}
    \item \textbf{Behavioural Cloning} -- in behavioural cloning, direct access to the underlying policy of the teacher agent is assumed, and used as a training signal.
    \item \textbf{Inverse Reinforcement Learning (IRL)} -- in IRL, the reward function is generated from the teacher's policy and then used to train the student model by existing reinforcement learning algorithms.
    \item \textbf{Observational Imitation Learning (OIL)} -- Rather than assuming direct access to the teacher's policy, in observational imitation learning, the learner agent only has access to external observations of the teacher agent acting in the learning environment.
\end{itemize}
These imitation learning techniques been widely used in the development of embodied agents; For instance RT2~\cite{zitkovich2023rt}, an embodied foundation model for robotics, was created by collating and training on a dataset made from a mix of human demonstrations and question answering web data. This model was then pretrained through OIL (in the form of next token prediction). Similarly, AlphaStar~\cite{vinyals2019grandmaster} was also pretrained from imitating games of professional players, and Ramrakhya et al.~\shortcite{ramrakhya2022habitat} utilise an embodied object-search strategy learnt from human demonstrations and robot trajectories.

\paragraph{Section Summary:}
Imitative Learning techniques are widely used, including in RL and NLP. There is nevertheless valuable work to be done exploring ways to increase the sample efficiency of these techniques, and whether socially embodied settings may play a role in this.

\section{Instructive Learning}
\subsection{Demonstration}
\label{sec: demo}
Demonstrations as an instructive learning paradigm consist largely in the provision of examples that a student agent can imitate. Importantly this means that the teacher may impart additional intent onto the demonstration through the choice and mode of presentation of those examples. For example a human teacher may draw attention to particular aspects of a demonstrated process by narrating it. Indeed human demonstrations are typically accompanied by explanations and narrations. Furthermore Ramrakhya et al.~\shortcite{ramrakhya2022habitat} verify that using imitation learning techniques on $70k$ human demonstrations outperforms using $240k$ agent trajectories, indicating that the quality of demonstrations can have a large impact. We have seen relatively little research exploring these elements of demonstration learning. Sumers et al.~\shortcite{sumers2023show} conduct a study on the provision of demonstrations to humans and find that if the demonstrations are not accompanied by explanations, they are less effective than linguistic interactions between teacher-student pairs for conveying concepts.

Bhoopchand et al.~\shortcite{bhoopchand2023learning} do explore some of the mechanisms at play with demonstrations, especially in the light of how they can be performed to maximise cultural transmission. They explore a navigation task with demonstrations performed by humans, that are then dropped out. Naive imitation (or rather emulation) of the experts enables the agent to perform well when the expert is there to follow, but once the expert is removed, these agents fail to generalise. The authors are able to create an agent that can imitate the plan of the teacher through equipping the agent with 1) memory, 2) an attentional bias towards the expert, 3) the expert ``dropping out'' during training, and 4) randomisation of the domain to ensure that the student agent is learning to utilise the expert and not just memorise the paths. This ``attentional bias'' may be related to the notion of intent as an aspect of ``true imitation''~\cite{tomasello1993cultural} and further accords well with Tomasello's~\shortcite{tomasello2009cultural} theory of joint attention.
\subsection{Feedback}
\label{sec: fed}
Feedback is just information given by the teacher on the performance of the learner. Consider the example of a student learning to perform addition. If they perform an incorrect addition like 18+17=25 we could provide a simple label evaluation (incorrect), a preferred answer (18+17=35), or language feedback like ``you forgot to carry over the tens term'' -- the particular form of feedback that is appropriate will depend on the situation.

Labels are of course omnipresent in machine learning, as supervised learning forms a major paradigm. Learning from preferences has also been developed especially in the field of RL, where \textit{Reinforcement Learning from Human Feedback} (RLHF) has now been adapted for finetuning language models~\cite{ouyang2022training}. While we mostly consider linguistic feedback, human feedback is often multimodal in nature, and may include gestures, tone, expressions and other physical actions. We encourage readers to refer to \cite{lin2020review} for progress in these dimensions; we expect these to become increasingly feasible and important as more works appear exploring multimodal models.

\subsubsection{Preference-Based Feedback}

Human preference-based feedback on generative AI output has proven invaluable for aligning the outputs of foundation models with human expectations through the use of `instruction tuning'~\cite{ouyang2022training}. This finetuning technique was designed for the purposes of \textit{alignment} -- the process of optimising a model according to human values and needs (e.g. following instructions). Typically this employs a three step process: 1) collecting demonstration data and performing supervised finetuning; 2) training a reward model on comparisons between model outputs; and 3) optimising the model as a policy against that reward model using an RL algorithm (PPO). The success of this approach has played an instrumental role in the recent rise of LLMs by providing a scalable and effective way to finetune models and control specific attributes.

Despite this, there have been a number of issues due to the cost of collecting human feedback data, and difficulty in optimising using PPO. One interesting development of late is Direct Preference Optimisation (DPO)~\cite{rafailov2024direct} -- which skips training the reward model and uses the LLM as an implicit reward model. This is essentially a KL-constrained contrastive loss that trains the model directly on preference datasets. Hejna et al.~\shortcite{hejna2024contrastive} point out that this method is not applicable to embodied trajectories and introduce a generalised version called ``contrastive preference optimisation'', wherein a dis-preffered trajectory is contrasted to a preferred one.

These developments should enable us to train embodied agents built with multimodal language models (e.g.~\cite{zitkovich2023rt}) using preference optimisation and bring this aspect of feedback into embodied agents.

\subsubsection{Natural Language-Based Feedback}
\label{sec: lbf}
While it is relatively simple to use comparisons or scalar feedback scores, fully leveraging the richness of language feedback has proven more of a challenge.

Sumers et al.~\shortcite{sumers2021learning} investigate learning under the regime of ``unconstrained linguistic feedback''. They categorise the feedback given as evaluative -- scalar feedback similar to that described above, imperative -- giving a preferred action e.g. ``you should have carried over the tens term'', or descriptive -- providing relevant information to the task. Sumers' approach to leveraging these rich rewards is to provide models that try to model the sentiment (a simple sentiment analysis model and a more complicated model handcrafted from Gricean pragmatics), as well as an inference network trained to predict the rewards given by the teacher. In their experiments, they firstly validate the importance of the more complex models that can leverage the richer language feedback, secondly they show that human teaching typically involves a mix of all three categories of feedback in human experiments.

More recently, several works have exploited the recent advances in NLP to use LLMs to directly update the reward functions used for training embodied models in accordance with the feedback. For example in Eureka~\cite{ma2023eureka} the authors train a robotic hand to execute complicated gestural object manipulation (spinning a pen around) by updating a reward function from observations. Similarly, Yu et al.~\shortcite{yu2023language} use LLMs to define reward parameters that are then optimised over for a variety of robot embodiments including a quadruped and arm based robot. They find that this approach is generalisable and achieves 90\% success rates across a wide variety of tasks. The import of such approaches is that they enable the utilisation of LLM's high-level cultural knowledge about actions to be translated into low-level actions.

A promising alternative solution to this is to use the feedback to condition future actions of the learning agent, and rely on existing learning mechanisms to learn from these actions. Scheurer et al.~\shortcite{scheurer2022learning} use this paradigm to learn from natural language feedback by generating multiple responses conditioned on feedback. The response that is semantically most similar to the feedback is then treated as the ``correct'' response and used to finetune the model. Similarly, Chen et al.~\shortcite{chen2024learning} explore ``Imitation learning from Language Feedback'' in which they let the learner refine its answers based on human feedback, and then finetune on these refinements.

Zha et al.~\shortcite{zha2023distilling} further explore the mechanisms of memorising and retrieving feedback for conditioning embodied agent actions on the fly. They create a system for distilling corrections into retrievable knowledge that can be used by the LLM-agent in future generations. They use the feedback to firstly perform replanning, and then to create a correction that can be placed in a knowledge base (rephrased to be ``contextually complete'') and retrieved in future situations where relevant. In particular, they use both visual and textual representations for the correction embedding so that it can be retrieved when a situation is similar in either dimension.

\subsection{Instruction and Explanation}
\label{sec: i&e}
Instruction following itself has been widely used as a task in the alignment of models. In particular, the RLHF framework described in Section~\ref{sec: lbf} was popularised in NLP for instruction following~\cite{ouyang2022training}. This alignment process (learning to follow instructions) has enabled foundation models to learn to perform complex tasks from instructions. A large body of works explore methods of instruction in the form of prompting techniques like chain of thought~\cite{wei2022chain}, that have been shown to improve reasoning abilities. For example, Mu et al.~\shortcite{mu2024embodiedgpt} finetune their embodied model on complicated instructions that require the use of chain of thought planning over egocentric videos.

Instruction following has also seen use as a paradigm in the popular benchmarks TEACh~\cite{padmakumar2022teach} and ALFRED~\cite{shridhar2020alfred}. TEACh consists in human-human dialogue sessions where one human is playing the role of a teacher or commander; the other acts as a student and must follow the instructions of the teacher, while being able to ask questions for clarification. ALFRED is a relatively simple task and involves following singular grounded instructions like ``walk to the coffee maker on the right''. The latter instruction can be seen as an example of the vision-language navigation task, which is currently one of the more popular paradigms in embodied AI, given that it was recently enabled by increased performance of LLMs. These benchmarks however do not focus on actually \textit{learning} from the interactions with the instructors per se, nor indeed are the instructions \textit{targeted} at the learning agent.

Saha et al.~\shortcite{saha2024can} explore more deeply modelling interactions between students and teachers using theory of mind. One of their research questions is the degree to which personalisations of explanations aid students. While they do find a positive effect, its influence is relatively small. While personalisation of explanations have not been found to be beneficial thus far, in an embodied RL setting it was found that the act of giving explanations itself was helpful in aiding the teacher learning~\cite{das2023state2explanation}. This is an example of the so called \textit{protégé effect} -- when we try to explain or instruct others we test and expand our own understanding of a concept by reconceptualising it.

While it would be desirable to heavily use targeted instructions and explanations by asking for help when needed, the provision of help, especially in the case of humans in the loop, is expensive. Indeed per the theory of scaffolding, we expect that as the learner gains proficiency, the learning experience can be less structured, and intervention can be reduced over time. As such one body of works considers the problem of efficiency by learning when to ask for help. For example Ren et al.~\shortcite{ren2023robots} train an agent to ask for help when there is a high level of uncertainty in the task. The agent generates a set of possible actions, and if the likelihood of the likeliest of them being correct (as measured by the agent) is below a certain threshold, the robot then asks for further instruction.

\subsection{Curricula}
\label{sec: cur}
Curriculum Learning is a machine learning approach that parallels the scaffolding of human learning at a more abstract level. The classic paradigm introduced by Bengio et al.~\shortcite{bengio2009curriculum} is to order the training data by complexity for some notion of complexity, and progressively increase the level of complexity as the learning progresses. Per Figure~\ref{fig:Learning Strategies}, curriculum learning consists in the teacher agent altering the learning environment rather than directly interacting with the learner agent.

The idea behind this is that by structuring the data in this way, the initial learning task becomes more feasible, and after basic concepts are acquired, these can be used as a basis for learning more complex concepts. While this seems simple, there are two key problems. First of all defining complexity for a given task is difficult -- the curriculum may have to be tailored for individual learners, and require insight into the current capabilities of the learner. Additionally it may be unclear when the learner has sufficiently mastered a concept to enable the complexity to increase.

Accordingly a body of works exists that try to automate this process. This approach is typically called automatic curriculum learning, and is a popular paradigm in the reinforcement learning community (ACL). For instance, Matiisen et al.~\shortcite{matiisen2019teacher} explore a ``Teacher Student Curriculum Learning'' paradigm in which a teacher agent monitors the student's training progress and is used to determine the tasks suitable for the student. The mechanism used is the student's progress on a task -- the teacher increases sample weight for tasks that are seeing progress -- the weight is then reduced when the task has been mastered since the rate of progress is measurably decreased. The authors treat the problem as a bandit learning problem and explore a number of sampling algorithms, finding that such an automated curriculum can help for maze solving in minecraft, and simple addition tasks using LSTMs.

Other works such as Ramrakhya et al.~\shortcite{ramrakhyal2023curriculum} and Morad et al.~\shortcite{morad2021embodied} use estimations of task success probabilities to order the experience while training embodied navigation systems. In particular, they aim to find tasks of `intermediate difficulty'. Interestingly, in the former they note that the addition of easy tasks to the training mix without curriculum learning per se has a similarly beneficial effect, as the learning agent simply spends less time in easier episodes since it can achieve them quickly.

Despite their success in RL domains,curriculum-based approaches have not seen as much success in NLP or computer vision. Recently Warstadt et al.~\shortcite{warstadt-etal-2023-findings} ran a competition to train LLMs using a human-developmentally plausible dataset and found that although curriculums were the most popular approach explored by entrants, it was relatively ineffective for improving sample efficiency. Indeed the key negative result of their competition report was exactly that none of the common curriculum learning approaches offered substantial benefits to the developmental task of language learning. Given the differing success of curriculum learning approaches it is likely that it's benefit is largely dependent on how structured the learning problem is.

Indeed \textit{embodied} language models may be a more suitable domain for the application of curriculum learning methods. Wang et al.~\shortcite{wang2024voyager} leverage the LLM's knowledge of task properties to derive new suitable tasks. Rather than relying on handcrafted curricula or designing a separate automatic curriculum, they simply prompt the LLM to come up with tasks itself, conditioning the new tasks on the state of the agent, previously learnt skills, the previous task completion history, and the long term goals of the agent. They find that this automatic curriculum is both critical for the agent's success, and furthermore beats out manually defined expert curricula. This process of giving the agent meta-learning abilities is promising but likely requires more principled analysis; in any case it speaks to the potential of using LLM-Agents to instantiate social learning paradigms in a unified way.

\paragraph{Section Summary:} More principled work remains to be done on demonstration learning, such as investigating whether ToM can enable student agents to imitate more efficiently. Feedback has been widely used in the form of preference learning. Leveraging general language feedback has seen increasing use but requires more research. Approaches like conditioning on feedback likely have similar applications to leveraging explanations and instructions. Finally, while curriculum learning has been widely used in RL, it is not as promising for training LLMs -- this may not however be the case for LLM-Agents.

\section{Collaborative Learning}
\subsection{Multi-Agent Reinforcement Learning (MARL)}
\label{sec: marl}
Historically the majority of research on collaborative learning has occurred in the context of multi-agent reinforcement learning. For instance AlphaStar~\cite{vinyals2019grandmaster} was trained to grandmaster human level in the game StarCraft II through playing with other players. The system was initialised using imitation learning, and then further finetuned by an RL algorithm to perform well against a mixture of opponents. The solution proposed by the authors is an evolution of previous so called ``self-play'' strategies. This consists in pitting different versions of an agent against itself in strategy games. The key insight here was to firstly add a greater mix of adversarial agents (previous versions of the learning agent) in addition to agents that were specifically designed to exploit the flaws of the agent in question.

Recently, an AI agent (Cicero) was trained for the game Diplomacy at a human level, ranking in the top $10\%$ of participants in an online league~\cite{meta2022human}. In Diplomacy, players take the role of diplomats controlling countries, and have to negotiate with other players, make promises, and deceive them. By leveraging pretrained LLMs to conduct the negotiation and model other users' intents, a strategic reasoning model could then be used to plan actions and intents and was then trained with a combination of RL methods including self-play and behavioural cloning.

In the more cooperative setting, a number of recent works have focused on the ``learn to communicate'' paradigm~\cite{oroojlooy2023review}. In these paradigms, the sets of agents may have to learn both what and when to communicate, and even some explorations of who to communicate to via attention mechanisms~\cite{gronauer2022multi}. Many of these works have started from a point of zero linguistic competence and require the agents to learn from scratch. Accordingly a future trend will likely include the incorporation of existing language models~\cite{gronauer2022multi}, similarly to the approach taken in the work of FAIR et al.~\shortcite{meta2022human} described above.

\subsection{Collaborative LLM-Agents}
\label{sec: llm col}
A significant recent trend in NLP has been exploring the use of communicative LLM-agents. Firstly, Park et al.~\shortcite{park2023generative} explored using such generative agents as models of human behaviour -- in particular, they simulated a town of agents that perform daily activities and interact in natural language, enabled by a game environment that tracks the memories and positions of each agent. 
These behaviours, including emergent abilities at party planning, were rated as being largely believable as human by crowdworkers.

Later, Qian et al.~\shortcite{qian2023communicative} explored the creation of software using teams of LLMs mimicking the `waterfall' model of software development. The authors broke down the development process into a sequence of discussions between LLM agents acting in roles (e.g. CEO, programmer etc.). In each discussion the LLMs would act in turn as instructors and assistants, making decisions through concensus. In contrast to simply generating the codebase for a software development project directly using an LLM designed for coding, they managed to achieve an $\sim$88\% success rate on achieving the software development task.

Chen et al.~\shortcite{chen2024agentverse} then went further to provide a more general framework for communicative LLM based agents. The process proceeds in rounds made up of 1) expert recruitment -- building up a group of roles to perform, 2) decision making -- deliberating over what to do 3) action execution, and 4) evaluation in which the outcome is compared to the goal and feedback is given.

In the embodied setting, Zhang et al.~\shortcite{zhang2024building} have explored building cooperative embodied LLM agents. This reportedly requires more complicated models and theory of mind - in particular their agents consist in modules handling belief, communication, reasoning, and planning. The belief module is intended to give the agents theory of mind by tracking the knowledge of the scene, other agents, the self, and the task. On benchmarks, the authors found that the LLM agents were able to cooperate well on various benchmarks, and that the theory of mind was necessary for this success.

Such cooperative LLM agents are already able to outperform state-of-the-art MARL agents on several tasks, such as navigation~\cite{li2023theory}, without the extensive number of training episodes typically required for the MARL agents. Importantly however, in all these collaborative settings, no learning per se is going on -- the agents are able to achieve complicated tasks through collaboration but only ``learn'' by updating the memory or context used in their internal decision making processes (since they are frozen language models).

More generally, we expect that collaborative settings are more useful for generating experience trajectories that would be unachievable for singular agents which can then be leveraged for learning using ordinary learning mechanisms, rather than being a source for learning mechanisms per se, in a similar way to Scheurer et al.~\shortcite{scheurer2022learning} exploring conditioning on feedback as a way of learning from natural language feedback rather than providing a separate feedback learning mechanism per se.

\paragraph{Section Summary:} Collaborative Learning has been explored mostly through Multi Agent Reinforcement Learning. These have included interesting studies on optimising the population of collaborators and emergent communication. In the case of LLM-Agents, there have been remarkable achievements through applying prompting, including application of ToM, but thus far limited work directly updating the weights of the models constituting these agents, limiting their adaptability.

\begin{table}[t]
    \centering
    \begin{tabular}{p{40mm}p{3mm}p{3mm}p{3mm}p{3mm}p{3mm}}
\toprule
                          Reference &   Iml &   Inl &  Col &  Emb &  ToM \\
\midrule
    \cite{belkhale2024rt} & Y & F,I & -- & Y & --\\
    \cite{chen2024agentverse} & -- & I,F & Y & -- & --\\
    \cite{chen2024learning} & Y & I,F & -- & -- & -- \\
    \cite{hejna2024contrastive} & -- & F & -- & Y & --\\
    \cite{mu2024embodiedgpt} & Y & I & -- & Y & --\\
    \cite{rafailov2024direct} & -- & F & -- & -- & --\\
    \cite{saha2024can} & Y & D,F & Y & -- & Y\\
    \cite{wang2024voyager} & -- & C,F & -- & Y & --\\
\cite{bhoopchand2023learning} & Y & D & -- & Y & --\\
\cite{das2023state2explanation} & -- & I & Y & Y & --\\
\cite{liu2023training} & Y & F,I & Y & -- & --\\
\cite{li2023theory} & -- & F & Y & -- & Y\\
\cite{ma2023eureka} & -- & F & -- & Y & --\\
\cite{park2023generative} & -- & & Y & Y & Y\\
\cite{qian2023communicative} & -- & F,I & -- & Y & --\\

\cite{ramrakhyal2023curriculum} & -- & C & -- & Y & --\\
\cite{ren2023robots} & -- & F & Y & Y & --\\
\cite{wang2023jarvis} & -- & I,F & -- & Y & --\\
\cite{yu2023language} & -- & F & -- & Y & --\\
\cite{zha2023distilling} & -- & F,I & -- & Y & --\\
\cite{zitkovich2023rt} & Y & D & -- & Y & --\\
\cite{meta2022human} & Y & & Y & Y & Y\\
\cite{ouyang2022training} & Y & I,F & -- & -- & --\\
\cite{padmakumar2022teach} & -- & I & Y & Y & --\\
\cite{ramrakhya2022habitat} & Y & D & -- & Y & --\\
\cite{scheurer2022learning} & Y & F,I & Y & -- & --\\
\cite{morad2021embodied} & -- & C & -- & Y & --\\
\cite{sumers2021learning} & -- & F & -- & Y & --\\
\cite{shridhar2020alfred} & -- & I & Y & Y & --\\
\cite{vinyals2019grandmaster} & Y & F,D & Y & Y & --\\
\cite{matiisen2019teacher} & -- & C & -- & Y & --\\
\cite{bengio2009curriculum} & -- & C & -- & -- & --\\
\bottomrule
\end{tabular}
    \caption{Social Learning techniques used: ImL: Imitative Learning, InL: Instructional Learning, CoL: Collaborative Learning, Emb: Embodied, ToM: Theory of Mind, C: Curriculum, D: Demonstrations, F: Feedback, I: Instruction, Y: Yes.} 
    \label{tab:booktabs}
\end{table}
\section{Conclusions and Future Directions}
Although social learning is not typically conceived as a key paradigm or organising principle for machine learning, we have shown that it permeates a large proportion of the machine learning literature.

Moreover, we have seen how performant LLMs have enabled the application of more advanced social learning techniques using the preferred mode of human instruction -- natural language. For example, in Section~\ref{sec: lbf} we showed that this has enabled the construction of models that can directly respond to language feedback, and in Section~\ref{sec: llm col} how they had been used to build communicative agents purely from language models that were able to outperform existing MARL approaches.

\paragraph{Comparison and Combination of Social Learning Paradigms.}
As shown in Table~\ref{tab:booktabs}, the majority of approaches explored in this paper use multiple social learning techniques. However, only four use all three of the main categories and are largely not embodied. As we have argued, equipping agents with linguistic capabilities has enabled agents to leverage communication and instructive learning paradigms out of the box. Given this, we expect to see more work exploring the combination of these approaches.

\paragraph{Learning Permanently through Social Conditioning.}
The promising LLM-Agent approaches have thus far largely relied on the knowledge acquired by the LLM through imitation learning. There has been little work performing further gradient updates once these models are socially embodied, and the ``learning'' has largely constituted updates to the agent's memory modules~(as in \cite{wang2024voyager,zha2023distilling}). Accordingly, identifying how models can learn more permanently in these settings remains a research priority. We see a promising direction in utilising low-level learning methods on experiences conditioned on the social interactions that entail ``higher-level'' learning mechanisms like feedback and instruction. This has seen some application in the non-embodied setting~\cite{scheurer2022learning}, but we expect this to be replicated for embodied settings in the near future as we move beyond simply using LLM-agents without finetuning.
\paragraph{Multimodality.}
Our survey has been primarily concerned with linguistic approaches, however the recent development of multimodal models will likely enable richer forms of social interaction (and thereby learning). We have already seen this in the case of demonstration learning from vision and language~\cite{zitkovich2023rt}, but given the relative immaturity of video-foundation models it may be some time before we are able to fully leverage non-linguistic signals like gestures and visual demonstrations for teaching embodied agents. As discussed in Section~\ref{sec: fed}, existing techniques for learning from feedback could be applied to learn from these signals.
\paragraph{Theory of Mind and Intent.}
As can be seen in Table~\ref{tab:booktabs}, there have been a number of approaches exploring the use of Theory of Mind as a prerequisite for richer interactions between teachers, students, and peers. Indeed Theory of Mind cuts across social learning. Modelling the intent of a teacher enables a learner agent to fully leverage the learning signal, as we discuss in Section~\ref{sec: demo}. Modelling teacher agents also allows learners to evaluate the teacher's potential as sources of learning~\cite{gweon2021inferential}. While ToM modules are a relatively common aspect of agents, especially collaborative agents, we have seen little work exploring these techniques so far. As such, we expect to see more work investigating the effects of these modules, especially given the interest in ToM initiated by Kosinski~\shortcite{kosinski2023theory}.

\paragraph{Application Domains.}
Social learning techniques are broadly applicable in the field at large. Nevertheless, they may play an especially important role in domains where social skills themselves are more relevant. For instance, in the case of socially-assistive robotics, the ability to adapt to individual needs and build a relationship over time is especially important~\cite{tapus2007socially}. Similarly, the ability to use models to teach humans skills would be a significant benefit of developing instructive social learning techniques. For example, this has explored in the case of social skill training~\cite{yang2024social}. In the meantime, we expect the majority of progress to occur in using social-learning techniques for training agents in video games, due to the low cost of training, ample amounts of training data, and ease of interacting with humans. Games are already a test-bed for many approaches covered above~\cite{wang2024voyager,wang2023jarvis,vinyals2019grandmaster}, but we can expect social learning to enable richer forms of collaborative play with AI agents, as explored for example in the work of Gong et al.~\shortcite{gong2023mindagent}.

\appendix
\section*{Acknowledgments}
We would like to thank Chloe He for her invaluable suggestions. This work was supported by an A*STAR CRF award to C.T., as well as an A*STAR SINGA scholarship to D.H.

\bibliographystyle{named}
\bibliography{references}

\end{document}